\documentclass[11pt,a4paper]{article}
\usepackage[hyperref]{acl2021}
\usepackage{times}
\usepackage{latexsym}

\usepackage{microtype}

\aclfinalcopy 
\def\aclpaperid{1362} 

\usepackage[T1]{fontenc}
\usepackage{type1cm}
\usepackage[american]{babel}
\usepackage{url}
\usepackage{amssymb} 
\usepackage{amsmath}
\usepackage{multirow} 
\usepackage{xspace}
\DeclareMathAlphabet{\mathcalbf}{OMS}{pzc}{b}{n}
\usepackage{microtype}

\usepackage[pdftex]{graphicx}
\usepackage{tabularx}
\usepackage{booktabs}
\usepackage{subcaption}
\DeclareGraphicsExtensions{.pdf,.ai,.jpg,.png}
\setkeys{Gin}{pagebox=artbox}

\graphicspath{{./}}

\newcommand{\bsfigure}[3][]{%
	\begin{figure}[t]
		\centering
		\includegraphics[#1]{#2}
		\caption{#3}\label{#2}%
	\end{figure}
}

\RequirePackage{type1cm}
\RequirePackage{color}
\RequirePackage{soul}
\setstcolor{blue}
\definecolor{lightgray}{rgb}{0.95,0.95,0.95}
\definecolor{lightgreen}{rgb}{0.56,0.93,0.56}
\definecolor{lightblue}{rgb}{0.3,0.3,0.9}
\definecolor{tgray}{rgb}{0.5,0.5,0.5}

\newsavebox\bscombox
\newcommand{\bscom}[3][]{%
	\sbox{\bscombox}{\fontsize{8}{9}\selectfont#1#2#3}
	\noindent
	\st{#2}{\selectfont
		\color{blue}#3\ifx\\#1\\\else{\fontsize{8}{9}\selectfont\color{violet}[#1]}\fi
	}
}

\begin{document}
	\title{Argument Undermining: Counter-Argument Generation by Attacking Weak Premises}

\newcommand{\lei}{\textsuperscript{$\ddagger$}}
\newcommand{\pb}{\textsuperscript{$\dagger$}}

\author{%
	Milad Alshomary \pb
	\qquad Shahbaz Syed \lei
	\qquad Arkajit Dhar \pb \\[1.5ex]
	\bfseries Martin Potthast \lei \hspace{1.5ex}
	\bfseries Henning Wachsmuth \pb \\[1.5ex] 
	\pb Paderborn University, Paderborn, Germany,   {\tt milad.alshomary@upb.de}\\
	\lei Leipzig University, Leipzig, Germany,  {\tt <first>.<last>@uni-leipzig.de} \\
}

\date{}

\maketitle

\begin{abstract}
Text generation has received a lot of attention in computational argumentation research as of recent. A particularly challenging task is the generation of {\em counter}-arguments. So far, approaches primarily focus on rebutting a given conclusion, yet other ways to counter an argument exist. In this work, we go beyond previous research by exploring argument {\em undermining}, that is, countering an argument by attacking one of its premises. We hypothesize that identifying the argument's weak premises is key to effective countering. Accordingly, we propose a pipeline approach that first assesses the premises' strength and then generates a counter-argument targeting the weak ones. On the one hand, both manual and automatic evaluation proves the importance of identifying weak premises in counter-argument generation. On the other hand, when considering {\em correctness} and {\em content richness}, human annotators favored our approach over state-of-the-art counter-argument generation.
\end{abstract}
	\section{Introduction}

Following \newcite{walton:2009}, a counter-argument can be defined as an attack on a specific argument by arguing against either its claim (called {\em rebuttal}), the validity of reasoning of its premises toward its claim ({\em undercut}), or the validity of one of its premises ({\em undermining}). Not only the mining and retrieval of counter-arguments have been studied \cite{peldszus:2015,wachsmuth:2018a}, recent works also tackled the generation of counter-arguments. Among these, \newcite{bilu:2015} and \newcite{hidey:2019} studied the task of contrastive claim generation, the former in a partly rule-based manner, the latter data-driven. Moreover, \newcite{hua:2019} proposed a neural approach. So far, however, research focused only on rebutting a given argument, ignoring the other aforementioned types. We expand this research by studying to what extent argument undermining can be utilized in counter-argument generation.

\begin{table}[t!]%
\centering%
\small
\renewcommand{\arraystretch}{1.0}
\setlength{\tabcolsep}{2.5pt}%
\begin{tabular}{p{0.9\columnwidth}}
\toprule
{\bf Claim (title):} Feminism is in the third wave. In countries, such as America, it has caused nothing but trouble. \\
\midrule
{\bf Premises (sentences):} I'm going to be bringing up several feminist arguments I have heard myself. First off, the all-dreaded wage gap. It has in fact been illegal to pay women less than men since the early 1960s. [...] Secondly the pink tax. Women's products are of course going to cost more than men's. They use entirely different chemicals specifically made to cater to softer skin [...] and for \textcolor{lightblue}{\em the fourth point, although I could go on for much longer, the feminst movement is needed elsewhere. In countries, such as Iraq, India, and Saudia Arabia. The feminst movement being in a country where women aren't being forced to cover their entire bodies, aren't being sold off with doweries, and aren't being oppressed, is downright absurd.} \\
\midrule
{\bf Counter-argument:} The fact that other women have it worse doesn't mean that women don't have it bad elsewhere. For example, I can be fired for being gay in 29 out of 50 states in the US. The fact that people are stoned for being gay in Brunei doesn't mean that isn't an example of homophobia...\\
\bottomrule
\end{tabular} 
\caption{An example argument (claim + premises) and a counter-argument in response to it, taken from Reddit changemyview. The italicized premise part was quoted by the user who stated the counter-argument.}
\label{table-intro-example}
\end{table}

In argument undermining, the validity of some premises is questioned. Such a phenomenon can be observed often in online discussions on social media. For example, in the discussion excerpt in Table~\ref{table-intro-example}, taken from the Reddit forum {\em changemyview},%
\footnote{https://en.wikipedia.org/wiki/R/changemyview}
a user contests the whole stated argument (claim and premises) refers to the specific premise highlighted in the table (on Reddit, it is the quoted part of the text). This implies two steps: first, to identify a potentially weak and thus attackable premise in the argument, and second, to counter it.

\enlargethispage{\baselineskip}
In this work, we propose to tackle the task of counter-argument generation by attacking one of the weak premises of an argument. We hypothesize that identifying a weak premise is key to effective counter-argument generation---especially when the argument is of high complexity, comprising multiple interlinked claims and premises, making it hard to comprehend the argument as a single unit. Figure~\ref{approach.pdf} illustrates our two-step pipeline approach: it detects premises that may be attackable and then generates a counter-argument addressing one or more of these premises. To identify weak premises, we build on the work of \newcite{jo:2020}, who classify attackable sentences using BERT. In contrast, we rank premises based on their attackability concerning the argument's main claim, utilizing the learning-to-rank approach of \newcite{han:2020}. For the second step, similar to \newcite{wolf:2019}, we fine-tune a pre-trained transformer-based language model \cite{radford:2018}, in a multi-task learning setting: next-token classification and counter-argument classification.

\bsfigure{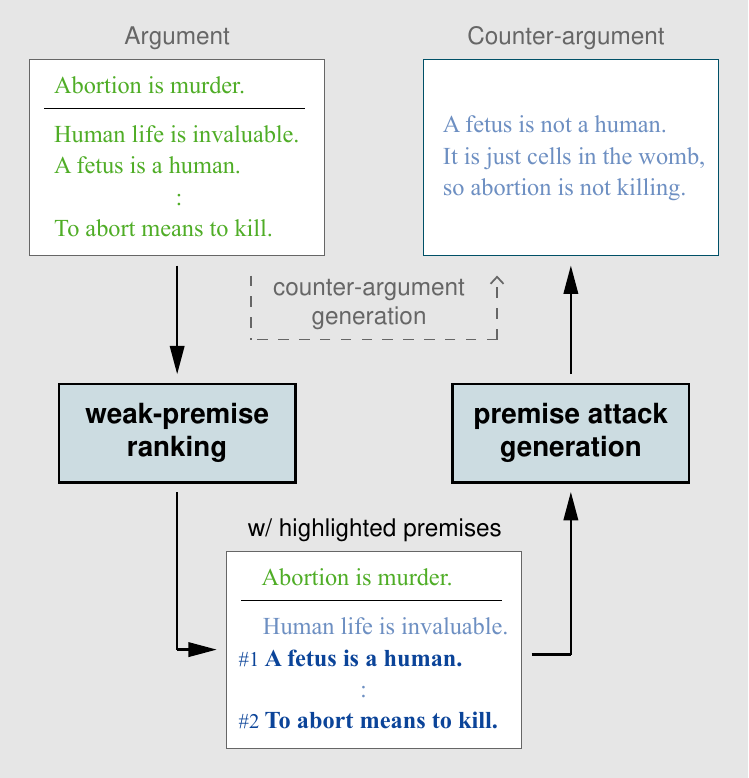}{Argument undermining: Instead of countering a given argument directly, our approach first ranks the argument's premises by weakness. Then, an attack focused on the weakest premises is generated.}

In our experiments, we make use of the changemyview (CMV) dataset of \newcite{jo:2020}, where each instance is a post consisting of a title (say, an argument's claim) and a text (the argument's premises). Some of the sentences in the text are quoted by comments to the post. These sentences are considered to be weak/attackable premises\footnote{Our assumption is that each sentence represents a premise supporting the main claim mentioned in the title of the post}. We further extend the dataset by collecting texts from comments, defining counter-arguments.

To analyze our approach, we evaluate both of its steps individually as well as in combination. In particular, we first compare our ranking model for detecting attackable premises to \newcite{jo:2020}, observing significant improvements in the effectiveness. Second, given the ground-truth attackable premise (the quoted sentences), we evaluate our counter-argument generation model against several baselines. Our automatic evaluation provides evidence that training the model with the weak premise annotated significantly boosts the scores across all metrics. We additionally confirm these results by a manual evaluation, indicating that our approach is better than the baseline in 56\% of the cases. Finally, we apply our generation model based on the automatically detected weak premises and compare it to the approach of \newcite{hua:2019}, which generates counter-arguments with opposing stance to the argument (i.e., rebuttals). While the automatic evaluation here was not in favor of our approach, the manual evaluation gave evidence of the favorability of our approach on all three tested quality dimensions. To summarize, our contributions are\footnote{Code and resources can be found \url{https://github.com/webis-de/ACL-21}}:
\begin{itemize}
	\setlength{\itemsep}{0pt}
	\item
	A model for detecting premise attackability, achieving state-of-the-art effectiveness.
	\item
	A new approach to counter-argument generation that identifies and attacks weak premises.
	\item
	Empirical evidence of the importance of considering a specific attackable premise in the argument when generating a counter-argument.
\end{itemize}
	\section{Related Work}

Recently, text generation has gained much interest in computational argumentation, both for single claims and complete arguments. \newcite{bilu:2015} composed opposing claims combining rules with classifiers, whereas \newcite{hidey:2019} tackled an analog task with neural methods. \newcite{alshomary:2020} reconstructed implicit claims from argument premises using triplet neural networks, and \newcite{gretz:2020} explored ability of GPT-2 to generate claims on topics. Recently, \newcite{alshomary:2021} studied how to encode specific beliefs into generated claims. \newcite{sato:2015} generated full arguments in a largely rule-based way. \newcite{el-baff:2019} modeled argument synthesis as a language modeling task, and \nopagebreak{\cite{schiller:2020}} studied the neural generation of arguments on a topic with controlled aspects and stance. Unlike all these, we deal with {\em counter}-arguments.

Research exists for mining attack relations \cite{cocarascu:2017,chakrabarty:2019,orbach:2020}, mining counter-considerations from text \cite{peldszus:2015}, and retrieving counter-arguments \cite{wachsmuth:2018a, orbach:2019}. However, only the consecutive works of \newcite{hua:2018,hua:2019} addressed the generation task. Their latest neural approach takes an argument or claim as input and generates a counter-argument rebutting it. Differently, we consider countering an argument by attacking one of its premises, known as {\em undermining} \cite{walton:2009}.

Part of our approach is to identify attackable premises, which can be studied from an argument quality perspective. That is, a premise is attackable when it lacks specific quality criteria. A significant body of research has studied argument quality assessment, with a comprehensive survey of quality criteria presented in \newcite{wachsmuth:2017a}. Implicitly, we target criteria such as a premise's acceptability or relevance. Still, we follow \newcite{jo:2020} in deriving attackability from the sentences of posts that users in the Reddit forum CMV attack. These sentences represent premises supporting the claim encoded in the post's title. The authors experimented with different features that potentially reflect weaknesses in the premises. Their best model for identifying attackable premises is a BERT-based classifier. We use their data to learn weak premise identification, but we address it as a learning-to-rank task.

As for text generation, significant advances have been made through fine-tuning large pre-trained language models \cite{solaiman:2019} on target tasks. We also benefit from this by utilizing a pre-trained transformer-based language model \cite{devlin:2018}, and we fine-tune it in a multi-task fashion similar to \newcite{wolf:2019}.
	\section{Approach}
\label{sec:approach}

As sketched in Figure ~\ref{approach.pdf}, our pipeline approach counters an argument by attacking the validity of one of its potentially weak premises. This section presents the two main steps of our approach: first, the ranking of weak and thus attackable premises, and second, the generation of the weak premises' attack.

\subsection{Weak-Premise Ranking}

Given an argument in the form of a claim and a set of premises, the task is to identify the argument's attackable premises. Unlike previous work \cite{jo:2020}, we model the task as a ranking task instead of a classification task, in which, for each argument, we learn to rank its premises by their weakness relevant to the claim. Our hypothesis here is that the attackability of a premise can be better learned when considering both the claim and other premises of the argument.

We operationalize the weak-premise ranking similar to the ranking approach of \newcite{han:2020}. In particular, given a set of premises and the claim, we first represent each premise by concatenating its tokens with the claim's tokens, separated by special tokens {\em [cls]} and {\em [sep]}:

\begin{center}
	\em
	[cls] claim\_tokens [sep] premise\_tokens [sep] 
\end{center}

\smallskip
Next, the resulting sequences are passed through a BERT model to obtain a vector representation for every premise. Each vector is then projected through a dense layer to get a score~$\hat{y}$ that reflects the weakness of the premise. Finally, a list-wise objective function (we use a Softmax loss) is optimized jointly on all premises of an argument as follows:
$$
l(y, \hat{y}) \;=\; - \sum_{i=1}^{n} y_i \cdot \log\Big(\frac{exp(\hat{y_i})}{\sum_{j=1}^{n} exp(\hat{y_{j}})}\Big),
$$

\noindent
where $y$ is a binary ground-truth label reflecting whether the given premise is attackable ($y=1$) or not ($y=0$). Given training data, we can thus learn to rank premises by weakness.

\subsection{Premise Attack Generation}

Given the ranking step's output, we identify the $k$ highest-ranked premises in an argument to be attackable (in our experiments, we test $k = 1$ and $k =3$). Then we generate a counter-argument putting the identified attackable premises into the focus. To this end, we follow \newcite{wolf:2019} in using transfer learning and fine-tune a pre-trained transformer-based generation model on our task.

\bsfigure{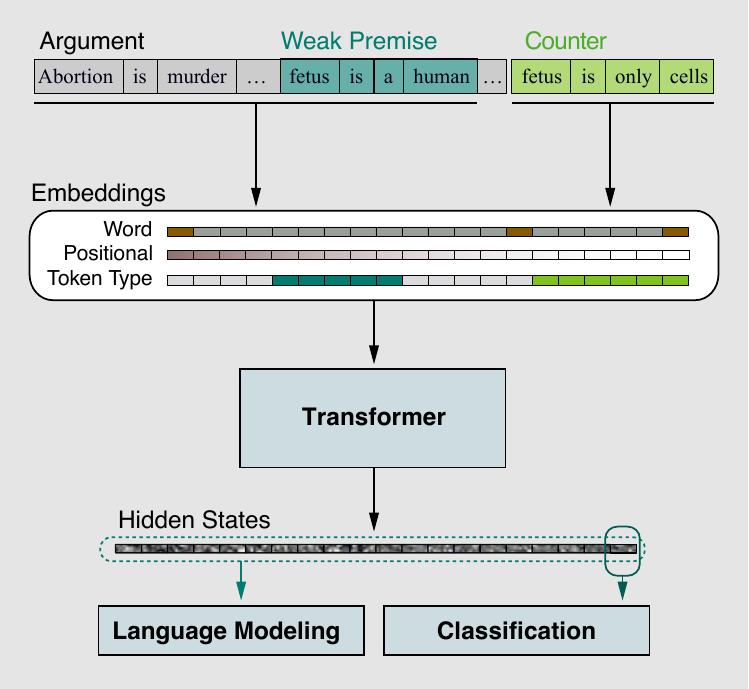}{Architecture of our approach: Given an argument, a weak premise, and a counter, three embedding representations are generated and fed to the transformer to obtain hidden states from which the language model and classification heads learn the {\em Next-token prediction} and {\em Counter-argument classification} tasks respectively}

In our fine-tuning process, the input is a sequence of tokens created from two segments, the argument and the counter-argument:%

\begin{center}
	\em
	[bos] arg\_tokens [counter] counter\_tokens [eos]
\end{center}

The final token embedding is then a result of concatenating three embeddings: word and positional embeddings learned in the pre-training process, as well as a token-type embedding learned in the fine-tuning process. Here, the token type reflects whether the token belongs to the argument in general, to a weak premise, or the counter-argument. Now, we train our model jointly on two tasks:
\begin{itemize}
	\setlength{\itemsep}{0pt}
	\item
	{\em Next-token prediction.} Given a sequence of tokens, predict the next one.
	\item
	{\em Counter-argument classification.} Given two concatenated segments, is the second a counter-argument to the first.
\end{itemize}

The first task is similar to the next-sentence prediction task introduced in \cite{devlin:2018}, which was shown to be beneficial for multiple representation-learning tasks. 

Figure~\ref{gpt-based-ca-gen.pdf} shows the architecture of our generation model. For training, we augment a given set of training sequences~$D$ by adding distracting sequences, which we use, for each argument and its weak premise, a non-relevant text instead of the counter-argument. Given a sequence of tokens $d = (t_1, t_2, \cdots, t_n) \in D$, we then optimize the following two loss functions jointly with equal weighting:
\begin{eqnarray*}
	L_1(\Theta) & = & \sum_{d \in D} \sum_{t_i \in d}^{} \log P(t_i \,|\, t_{i-k}, \cdots,t_{i-1}; \Theta), \\
	L_2(\Theta) & = & \sum_{d_j \in D}^{} \log P(y_j \,|\, t_1, \cdots, t_n; \Theta),
\end{eqnarray*}

\noindent
where $\Theta$ denotes the weights of the model, $k$ is the number of previous tokens,  and $y_j$ is the ground-truth label of the sequence, indicating if the second segment of the sequence is a counter or not.
	\section{Data}

As proposed, the presented approach models the task of counter-argument generation as an attack on a potentially attackable premise. Such behavior is widely observed on the Reddit forum {\em changemyview (CMV)}. In particular, a user writes a new {\em post} that presents reasons supporting the pro or con stance towards a given topic (captured in the {\em title} of the post), asking the CMV community to challenge the presented view. In turn, other users quote specific segments of the post (usually a few sentences) and seek to counter them in their {\em comment}. An example has already been given in Figure \ref{table-intro-example}. 

The structure induced by CMV defines a suitable data source for our study. Specifically, we create the following distantly supervised mapping: 
\begin{itemize}
	\setlength{\itemsep}{0pt}
	\item
	The title of the post denotes the {\em claim} of the user's argument;
	\item
	the text of the post denotes the concatenated set of the argument's {\em premises};
	\item
	the quoted sentence(s) denote the {\em attackable (weak) premises}; and 
	\item
	the quoting sentences from the comment denote the {\em counter-argument}.
\end{itemize}

In our work, we build on the CMV dataset of \newcite{jo:2020}, where each instance contains a post, a title, and a set of attackable sentences (those quoted in the comments). We use the same split as the authors, consisting of 25.8k posts for training, 8.7k for validation, and 8.5k for testing. We extend their dataset by further collecting the quoting sentences from the comments (i.e., the counter-arguments). The final dataset compiles 111.9k triples of argument (claim and premises), weak premise (one sentence or more), and a counter-argument (a set of sentences), split into 67.6k training, 23k validation, and 22.3k for testing instances.
	\section{Evaluation}
\label{sec:evaluation}

In the following, we present the experiments we carried out to evaluate both steps of our approach individually as well as in a pipelined approach. On the one hand, we aim to assess the applicability of identifying weak premises in an argument and the impact of targeting them in the process of counter-argument generation. On the other hand, our goal is to assess how well counter-argument generation via {\em undermining} is compared to other known counter-argument generation approaches.

\subsection{Weak-Premise Ranking}

As presented, we approached the task of finding attackable premises by learning to rank premises by their weakness with respect to the main claim. 

\paragraph{Approach}

Based on the code of \newcite{han:2020} available in the Tensorflow learn-to-rank framework \cite{pasumarthi:2019}, we used a list-wise optimization technique that considers the order of all premises in the same argument.%
\footnote{We also experimented with point-wise and pairwise techniques, but list-wise was best. } 
We trained our ranking approach on the CMV dataset's training split and referred to it as {\em bert-ltr} below.\footnote{Training details can be found in the appendix.}

\paragraph{Baselines}

We compare our approach to the Bert-based classifier introduced by \newcite{jo:2020}, trained on the same training split using the authors' code. We use their trained model to score each premise and then rank all premises in an argument accordingly. We call this {\em bert-classifier}. As \newcite{jo:2020}, we also consider a {\em random baseline} as well as a baseline that ranks premises based on {\em sentence length}.

\paragraph{Measures}

To assess the effectiveness, we followed \newcite{jo:2020} in computing the precision of putting a weak premise in the first rank ($P@1$), as well as the accuracy of having at least a weak premise ranked in the top three ($A@3$).

\paragraph{Results}

Table \ref{weak-premise-eval-table} shows the weak-premise ranking results. We managed to almost exactly reproduce the values of \newcite{jo:2020} for all three baselines. Our approach, {\em bert-ltr}, achieves the best scores according to both measures. In terms of a one-tailed dependent student-$t$ test, the differences between {\em bert-ltr} and {\em bert-classifier} results are significant with at least 95\% confidence. These results support our hypothesis of the importance of tackling the task as a ranking task with respect to the main claim. Below, we will use our weak-premise ranking model in the overall approach, i.e., to automatically select attackable premises in an argument.

\begin{table}[t!]%
\centering%
\small
\renewcommand{\arraystretch}{1}
\setlength{\tabcolsep}{5pt}%
\begin{tabular}{lrr}
\toprule
\bf Approach & \bf P@1 & \bf A@3 \\
\midrule
Random & 0.425  & 0.738 \\
Sentence Length & 0.350 &  0.617 \\
bert-classifier \cite{jo:2020}& 0.487 &  0.777 \\
bert-ltr (our approach)& \bf \textsuperscript{*}0.506 & \bf \textsuperscript{*}0.786 \\ 
\bottomrule
\end{tabular}%
\caption{Weak-premise ranking: Precision of ranking a weak premise highest (P@1) and accuracy for the top three (A@3) of all evaluated approaches. Results with * are significantly better than {\em bert-classifier} at $p < .05$.}
\label{weak-premise-eval-table}%
\end{table}

\subsection{Premise Attack Generation}
\label{eval-weak-premise-ranking}

Next, we evaluate our hypothesis of the importance of identifying weak premises in the process of counter-argument generation. To focus on this step, we use the ground-truth weak premises in our data. These are the quoted sentences in the post, considered potentially attackable premises.

\paragraph{Approach}

We used OpenAI GPT as a pre-trained language model. We trained two versions of our generation model: {\em our-model-w/} with an extra special token (\textit{[weak]}), surrounding the attackable sentences to give an extra signal to our model, and once {\em our-model-w/o} without it. We fine-tuned both versions with the same settings using the transformers library \cite{wolf:2020} for six epochs\footnote{We stopped at six epochs because we observed no gain in terms of validation loss}. We left all other hyperparameters to their default values.  As mentioned, the model's input is a sequence of tokens constructed from the argument (with weak premises highlighted) and either the correct counter or a distracting sequence. We randomly select one sentence from the original post to be the distracting sequence for each input instance.

\begin{table*}[t!]%
\centering%
\small
\renewcommand{\arraystretch}{1}
\setlength{\tabcolsep}{4pt}%
\begin{tabular}{lllrrrrrr}
\toprule
& & & \multicolumn{3}{c}{\bf Counter Sentences} & \multicolumn{3}{c}{ \bf Full Comment} \\
 \cmidrule(l@{5pt}r@{5pt}){4-6} \cmidrule(l@{5pt}r@{5pt}){7-9}
\# &\bf Approach & \bf Target & \bf METEOR & \bf BLEU-1 & \bf BLEU-2  & \bf METEOR & \bf BLEU-1 & \bf BLEU-2\\
\midrule
1 & counter-baseline & - 			&     0.058 &  13.023  &  3.117 & 0.097 &  10.400  &  3.212 \\
\addlinespace
2 & our-model-w/o & claim 	& \bf 0.060&     12.532&  2.943&     0.090&     9.472& 2.837 \\
3 & our-model-w/o & random premise&     0.058&     12.838&  3.005&     0.096&     10.398& 3.255 \\
4 & our-model-w/o & weak premise  &     0.057& \bf \textsuperscript{*}13.453&  \bf \textsuperscript{*}3.391& \bf \textsuperscript{*}0.102& \bf \textsuperscript{*}10.998& \bf \textsuperscript{*}3.764 \\
\addlinespace
5 & our-model-w/   & claim	& \bf 0.060&     12.635&  3.023&     0.092&     9.685& 2.984 \\
6 & our-model-w/   & random premise&     0.059&     12.712&  2.987&     0.096&     10.161& 3.217 \\
7 & our-model-w/   & weak premise 	&     0.058&     13.162&  3.217&     0.101&     10.743& 3.651 \\
\bottomrule
\end{tabular}%
\caption{Premise attack generation: METEOR and BLEU scores of the output of each evaluated approach compared to the ground-truth counter sentences and {\em the full comment (argument)}. Values marked with * are significantly better than {\em counter-baseline} at $p < .05$.}
\label{counter-gen-eval-table}%
\end{table*}


\paragraph{Baseline}

We compare our model to a GPT-based model fine-tuned on a sequence of tokens representing a pair of an argument (title and post) and a counter-argument. We consider this as a general counter-argument generation model, trained without any consideration of weak premises. We train the baseline using the same setting as our model. We refer to it as {\em counter-baseline}.

\paragraph{Automatic Evaluation}

To assess the importance of selecting attackable sentences, we evaluated the effectiveness of our model in different inference settings in terms of what is being attacked: 
(1)~the {\em claim} of the argument, 
(2)~a {\em random premise}, or 
(3)~a {\em weak premise} given in the ground-truth data. 
In the random setting, we randomly selected three premises from the argument, and we generated one counter for each. The final result is the average of the results for each.

\bsfigure{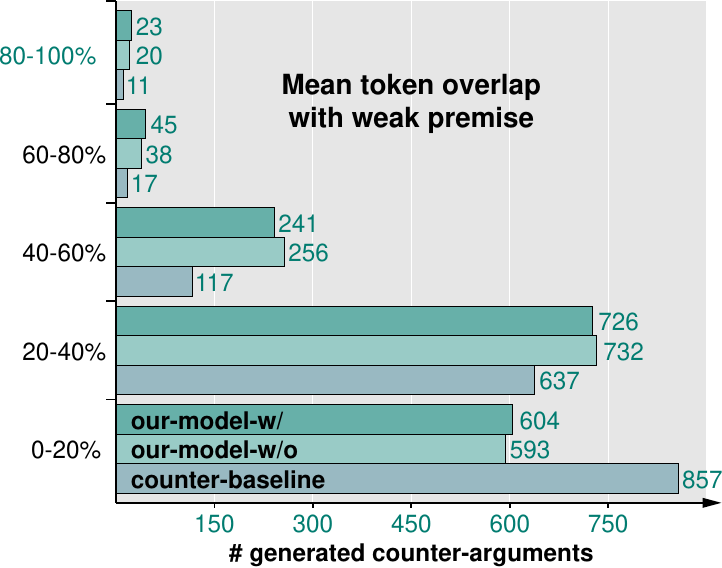}{Premise attack generation: Mean token overlap between the ground-truth weak premises and the counters generated by each evaluated approach.}

We computed METEOR and BLEU scores, comparing the generated premises to (a)~the exact counter sentences of the quoted weak premise and (b)~the full argument. We performed this automatic evaluation on 1k posts from the test split.

\paragraph{Results} 

As shown in Table \ref{counter-gen-eval-table}, the best results are achieved by {\em our-model-w/o} in all cases when identifying the weak premises in the input. Encoding the knowledge about weak premises as token types is sufficient, and adding an extra special token doesn't help. Although the differences between our best model and the baseline are not big, they are significant according to the one-tailed dependent student-$t$ test with a confidence level of 95\%. 
For both versions of our model, best scores are achieved when considering the weak premises as the target (except for the first METEOR column). However, not all these differences are significant. This gives evidence that exploiting information about weak premises in the training of counter-argument generation approaches can improve their effectiveness. 

To further assess the relationship between the generated counters and the attacked premises, we computed the proportion of covered content tokens in the weak premise for the two versions of our model and the baseline. Figure \ref{overlap-with-weak-premise.ai} shows a histogram of the percentages. Clearly, both versions of our model have higher coverage of the annotated weak premise than the baseline. 

\begin{table}[t!]%
	\centering%
	\small
	\renewcommand{\arraystretch}{1}
	\setlength{\tabcolsep}{4pt}%
	\begin{tabular}{lcccc}
		\toprule
		\bf Approach & \multicolumn{2}{c}{\bf Relevance} & \multicolumn{2}{c}{\bf Appropriateness} \\
		 \cmidrule(l@{2pt}r@{2pt}){2-3} \cmidrule(l@{2pt}r@{2pt}){4-5}
		& \bf Majority & \bf Full  & \bf Majority & \bf Full \\
		\midrule
		counter-baseline & 44\%  	& 20\%  	& 44\%		& 14\% \\ 
		our-model-w/o    & \bf 56\% & \bf 36\%  & \bf 56\%	& \bf 28\% \\
\midrule
Mean Kendall's $\tau$ & \multicolumn{2}{c}{0.41} & \multicolumn{2}{c}{0.23}  \\
		\bottomrule
	\end{tabular}%
	\caption{Premise attack generation: Percentage of cases where each given approach was seen as more relevant and more appropriate, respectively, according to majority vote and the full agreement in the manual evaluation on 50 examples. The bottom line shows the mean pairwise inter-annotator agreement.}
	\label{manual-eval-1-table}%
\end{table}

\paragraph{Manual Evaluation} 

To analyze the generated counter-arguments more thoroughly, we carried a manual evaluation study on a sample of 50 random examples. Two authors of the paper inspected the sample comparing the two versions of our model. The results were in favor of {\em our-model-w/o}. Therefore, we compared only {\em our-model-w/o} against the {\em counter-baseline}. In particular, we assessed the relevance and appropriateness of the output of the two for each example. Given an argument, the highlighted premise to be attacked, and the two counters, we asked three annotators who hold an academic degree and are fluent in English (no author of this paper) to answer two questions:
\begin{enumerate}
	\setlength{\itemsep}{0pt}
	\item
	Which text is more relevant to the highlighted premise?
	\item
	Which text is more appropriate for being used as a counter-argument?
\end{enumerate}

\paragraph{Results}

As shown in Table \ref{manual-eval-1-table}, considering the majority vote, annotators favored our model in 56\% of the cases in both tasks. These results give further evidence supporting our hypothesis of the importance of identifying weak premises. Considering the given task as a ranking task, we used Kendall's $\tau$ to compute the annotator's agreement. The mean pairwise agreement was  0.41 for the relevance assessment and 0.23 for appropriateness. Clearly, assessing the text's appropriateness of being a counter-argument is more subjective and more challenging to judge than the relevance task.

\subsection{Overall Approach}

Finally, we assess the overall effectiveness of our counter-generation approach, that is, when we identify weak premises automatically using {\em Bert-ltr} and then generating a counter-argument using  {\em our-model-w/o},  focusing on the selected weak premises. 

\paragraph{Approach}

Due to the limited {\em P@1} value of our ranking model (see Table \ref{weak-premise-eval-table}), we evaluate two variations of our overall approach that differ in terms of what premises to attack. The first variant attacks the weakest premise. In the second, we first generate three counters considering each of the top three weak premises. Then, we select the counter that has the most content-token overlap with the corresponding weak premise.

\begin{table}[t!]%
\centering%
\small
\renewcommand{\arraystretch}{1}
\setlength{\tabcolsep}{2pt}%
\begin{tabular}{lll@{\hspace*{-1em}}rrr}
\toprule
\# &\bf Approach & \bf Target & \bf METEOR & \bf BLEU-1 & \bf BLEU-2\\
\midrule
1 & counter-baseline & None 					&     0.205 &  22.741  &  7.792\\
2 & Hua and Wang	   & None		   &     \bf 0.258 &  \bf 30.160   &  \bf 13.366 \\
\addlinespace
3 & Overall approach             & 1 premise	&    0.207 &  22.841  &  7.839\\
4 & Overall approach 			  & 3 premises&     0.210 &  23.400  &  8.025 \\

\bottomrule
\end{tabular}%
\caption{Overall approach: METEOR and BLUE scores of the two variants with different attacked targets, the counter-baseline, and \newcite{hua:2019}.}
\label{overall-automatic-eval-table}%
\end{table}

\paragraph{Baselines}

On the one hand, we compare our approach to the {\em counter-baseline} from the previous section. On the other hand, we consider the state-of-the-art counter-argument generation approach of \newcite{hua:2019}, an LSTM-based Seq2seq model with two decoders, one for selecting talking points (phrases) and the other for generating the counter given the selection. 

\paragraph{Automatic Evaluation}

While the approach of \newcite{hua:2019} learns from a dataset collected from the same source (CMV), it requires retrieving relevant argumentative texts with a stance opposite to the input argument. Due to the complexity of the data preparation, we decided instead to evaluate all approaches on the test split of \newcite{hua:2019}.%
\footnote{We verified that all posts in their test split do not appear in our training split.}
As a result,  \newcite{hua:2019}'s approach is trained on their training split, while our approach is trained on our training split, and then both are evaluated on the same test split of \newcite{hua:2019}. This can be considered a somewhat unfair setting for our approach due to certain domain differences since the dataset of \newcite{hua:2019} comprises political topics only.
Similar to Section~\ref{eval-weak-premise-ranking}, we generated counters for 1k examples and computed METEOR and BLEU scores of the generated counters with respect to the ground-truth counters, which are here full arguments (CMV comments).

\paragraph{Results} 

Table \ref{overall-automatic-eval-table} shows that our approach outperforms the counter-baseline in both settings, even with weak premises selected automatically. Considering the top-3 weak premises instead of the top-1 improves the results. The best scores are achieved by {\em Hua and Wang}, though. A reason for this may be the slight domain difference between our model's training data and the test data used for evaluation. Another observation is that the scores of both our approach and the baseline increase compared to Table~\ref{counter-gen-eval-table}. This is likely to be caused by the higher number of ground-truth references for each instance in data of \newcite{hua:2019} compared to the test split of our data, making it more likely to have token overlaps.

\paragraph{Manual Evaluation} 

Given the known limited reliability of automatic generation evaluation, we conducted another user study to evaluate the quality of the generated counters by our model and {\em Hua and Wang}. We evaluate the same quality dimensions used previously by \newcite{hua:2019}:
\begin{itemize}
	\setlength{\itemsep}{0pt}
	\item
	{\em Content Richness.} The diversity of aspects covered by a counter-argument.
	\item
	{\em Correctness.} The relevance of a counter-argument to the given argument and their degree of disagreement.
	\item
	{\em Grammaticality.} The grammatical correctness and fluency of a counter-argument.
\end{itemize}

We used the Upwork crowd working platform to recruit three annotators with English proficiency and experience in editorial work.%
\footnote{Upwork, \url{http://upwork.com}}
We asked each of them to evaluate a sample of 100 examples. Each contained an argument (claim and premises) and two counters (one of each approach). We asked the annotators to compare the counters and assess each with a score from 1 (worst) to 5~(best) for each quality dimension.

\begin{table}[t!]%
\centering%
\small
\renewcommand{\arraystretch}{1}
\setlength{\tabcolsep}{4pt}%
\begin{tabular}{l@{\hspace*{-0.8em}}rrr}
\toprule
& {\bf Correctness} & {\bf Richness} & {\bf Grammaticality} \\
\midrule
Hua and Wang & 1.81 & 2.28 & 2.91\\
Overall approach & \bf 2.65 &\bf 3.15 &\bf 3.50\\ 
\midrule
Krippendorff's $\alpha$ & 0.26 & 0.06  & 0.32 \\
\bottomrule
\end{tabular}%
\caption{Overall approach: Average scores of the three annotators for the three evaluated quality dimensions of the counter-arguments generated by our approach and the one of \newcite{hua:2019}. 1 is worst, 5 is best. The bottom line shows the inter-annotator agreeement.}
\label{table-final-evaluation}%
\end{table}

\paragraph{Results}

The results are presented in Table \ref{table-final-evaluation}. Unlike in the automatic evaluation, the annotators gave, on average, higher scores on all quality dimensions to our generated counters than to those of {\em Hua and Wang}.%
\footnote{We note that the scores of Hua and Wang in Table \ref{table-final-evaluation} are notably lower than those reported by \newcite{hua:2019}. We believe this to be due to the comparison with our approach that affected the annotator's scores.}
Bringing knowledge from pre-trained language models (GPT) generally seems to contribute to the {\em grammaticality} and the {\em richness} of the generated counters. In terms of generating a {\em correct} counter, focusing the generation model on a specific weak premise in an argument seems to help (2.65 vs.\ 1.81), even though the results are far from perfect. Manual inspection revealed that far from all generated arguments are counters to exactly what is in the argument, indicating more room to work on this topic.

\begin{figure}[t!]%
	\centering%
	\small
	\renewcommand{\arraystretch}{1.0}
	\setlength{\tabcolsep}{2.5pt}%
	\begin{tabular}{p{\columnwidth}}
		\toprule
		{\bf Claim:} there's nothing wrong with income inequality. \\
		\midrule
		{\bf Premises:} billionaires like the rockefellers and trump worked hard to earn their money and provide their families with luxury. meanwhile, my grandpa didn't and because of that i am of lower-middle class status. \textcolor{lightblue}{\em just because i'm poorer doesn't mean i'm entitled to the cash that the rich spent years to accumulate}. i simply have to swallow my pride and start at the bottom . to try and stump income inequality is to meddle with the very basis of pure , unadulterated capitalism and meritocracy\\
		\midrule
		{\bf Our Counter:} income inequality is precisely because of the fact that rich people have a vested interest in having a bunch of money in the first place, which leads to great wealth inequality. if you are poor but poor you have no incentive to live. by providing good quality goods and services, you are able to contribute to society better than most poor people. you're basically doing something that is expected of you, which is a wonderful trait. i think the problem is that you should at least be able to be generous towards someone without expecting to be rewarded.
		 \\
		\midrule
		{\bf Hua and Wang's Counter:} this is a great example of how hard it is to invest in low income housing. it's not like it's going to end up being worse for everyone. if you don't like it, you're going to have to worry about it. the rich don't want to pay for it because they do n't have to pay taxes. they aren't going to be able to do anything about it, they just don't want.
		 \\
		\bottomrule
	\end{tabular} 
	\caption{Example counter-arguments generated by our approach and by the approach of \newcite{hua:2019}. The italicized premise segment was identified as the weak premise by our approach.}
	\label{table-example-generated-counters}
\end{figure}

Krippendorff's $\alpha$ values show that the annotators had a fair agreement on {\em grammaticallity} and {\em correctness} tasks (given the subjectiveness of the tasks), but only slight agreement on {\em content richness}. We, therefore, think that the results for the latter should not be overinterpreted.

In Figure \ref{table-example-generated-counters}, we show an example argument in favor of {\em income inequality}. Our approach considers the premise ``being poor does not entitle someone to the cash of the rich people''. It then generates a counter-argument on the topic of inequality, focusing on the fact that ``being poor limits the ability to contribute to society". In contrast, the counter-argument generated by {\em Hua and Wang} diverges to address ``low-income housing'' which is less relevant to the topic. More examples of generated counters can be found in Figure~5 in the appendix.
	\section{Conclusion}

In this work, we have proposed a new approach to counter-argument generation. The approach focuses on argument {\em undermining} rather than {\em rebuttal}, aiming to expand the research in this area. The underlying hypothesis is that identifying weak premises in an argument is essential for effective countering. 
To account for this hypothesis, our approach first ranks the argument's premises by weakness and then generates a counter-argument to attack the weakest ones. 

In our experiments, we have first evaluated each step individually. We have observed state-of-the-art results in the weak-premise identification task. Our results also support the need for identifying weak premises to generate better attacks. We have also evaluated the overall approach against the state-of-the-art approach of \newcite{hua:2019}. While we did not beat that approach in automatic evaluation scores, independent annotators favored the counter-arguments generated by our approach across all evaluated quality dimensions.

We conclude that our approach improves the state of the art in counter-argument generation in different respects, providing support for our hypothesis. Still, the limited manual evaluation scores imply notable room for improvement. Most importantly, controlling the stance of the generated counters is yet to be fully solved.

	\section{Ethical Statement}
We acknowledge that ethical issues might arise from our work. First, we would like to ensure that we did not violate user privacy when using data from public platforms. By reusing pre-trained models, our approach might have inherited some forms of bias. Mitigating such bias is still ongoing research. It is worth noting that our experiments show that our approach is far from being ready to be used as an end technology. Our goal is to advance the research on this task.

\section*{Acknowledgments}

This work was partially supported by the German Research Foundation (DFG) within the Collaborative Research Center ``On-The-Fly Computing'' (SFB~901/3) under the project number~160364472.

	\bibliography{acl21-argument-undermining-lit}
	\bibliographystyle{acl_natbib}
	
	\appendix
	\section{Models and Training Specifications}
All our models were trained on one {\em 24 GeForce GTX 1080 GPU}.

\paragraph{Weak-Premise Ranking}
We use the code of \cite{han:2020} to train our ranking model {\em Bert-ltr}, with BERT-Base, Uncased for embedding (12-layer, 768-hidden, 12-heads, 110M parameters). The model was trained for 100k steps, which took almost 9 hours of training.

\paragraph{Premise Attack Generation}
We use the transformers library to train our generation models, with GPT as a pre-training language model (12-layer, 768-hidden, 12-heads, 110M parameters. OpenAI GPT English model). The model was fine-tuned for six epochs, which took almost one day of training. In the generation time, we use top-$k$ sampling technique \cite{huang:2019}, with the following parameters: top-$k$=50, top-$p$=0.95, and temperature=1.0. We generate counters of a minimum of 100 tokens and a maximum of 150 tokens.

\paragraph{Evaluation Measures}
The BLEU and METEOR scores are computed using the NLTK library (\url{https://www.nltk.org/}). Significance tests were performed using the Scipy library (\url{https://www.scipy.org/}).

\section{Example Counters}
In this section, we present example counters generated by our approach and \newcite{hua:2019}

\begin{figure*}[t!]%
	\centering%
	\small
	\renewcommand{\arraystretch}{1.0}
	\setlength{\tabcolsep}{2.5pt}%
	\begin{tabular}{p{0.9\textwidth}}
		\toprule
		{\bf Claim (title):} it's better to have no freedom of speech than to be given the allusion of freedom of speech. \\
		\midrule
		{\bf Premises (sentences):} the reason why i believe this is that i like to know the rules first upfront , i hate people changing the rules in the middle of the game . the simplest comparison i can come up with is that if i am walking and then i accidentally walk into someone 's property than i get arrested for trespassing . i rather they straight up hang signs that say i am entering private property in the first place . \textcolor{lightblue}{\em if i go my whole life believing i have free speech , then one day i get maced or tasered by a cop for speaking my mind }. i would 've expressed my opinions in another way . these countries where they say you have freedom but on the other hand oppress you are not free countries nor are they honest countries . in my view , they are worse than an openly oppressive country . \\
		\midrule
		{\bf Our Counter:} can you expand on what you mean here? this is simply demonstrably false. we have free speech. people have free speech in these countries because we are free people. i'm not sure how you got from you that you dislike freedom, but there's no contradiction. i am not free of speech. \\
		\midrule
		{\bf Hua and Wang's Counter:} if you believe in the right to free speech , then it is not a right . if you do n't agree with what you say , then you have to accept that it is a slippery slope . it 's not that you are n't allowed to say whatever you want , it 's that you do not have the right , and that you should not be allowed to speak freely . you can not have a right to express your views , but that does n't mean you should n't do it . if you \\
		\bottomrule
		\toprule
		{\bf Claim (title):} i think hijab is good for the society and for women. \\
		\midrule
		{\bf Premises (sentences):} but i think that hijab is a good approach to reduce rape and harassment . \textcolor{lightblue}{\em i think one of the reasons for rape are 1 ) not having a sexual partner 2 ) pornography , so here comes the role of the hijab so if women wear hijab rapists will not view her as a sexual object} . here is a video that shows a social experiment of the same woman wearing different cloth on the same streets of nyc :[ -- url -- ) . i want people that view hijab as some sort of backward way of dress to tell me why do you think like that ? and if possible change my view . ( edit ) .... . i agree with many of you guys that said rape is about power , but what i meant in my question is sexual harassment in general , by that i mean touching or grabbing private parts i could n't explain at first because of my english . so if touching and grabbing is under the crime of sexual harassment , do n't you think hijab will reduce those assaults ? so in short my point of view has changed when you guys said that rape is about power , i know think that a rapist will rape anyone regardless his cloth , but that does n't s include other sexual harassment means . excuse my english .\\
		\midrule
		{\bf Our Counter:} this point you're making is'why do you think women are raped if you'd like to do x with them?'i'm not sure what you're trying to get at here. are you saying people shouldn't think rape is a'bad idea '? because that's the most basic and fundamental reason a person should wear a hijab, and unless you don't believe in the concept of a'good idea ', you shouldn't think women are sexually harassed. rape is, in general, about power, sexual harassment, by the very definition of harassment.  \\
		\midrule
		{\bf Hua and Wang's Counter:} there is no such thing as a " feminist " movement . it 's not just a matter of what it is , it 's a group of people who are in favor of it . women are not the only ones who are oppressed . women have been subject to the same rules as men . women are
		 \\
		\bottomrule
		\toprule
		{\bf Claim (title):} there is no logical way to prove that being unfair is wrong .\\
		\midrule
		{\bf Premises (sentences):} . one response was that " the party did n't deserve that type of treatment . " but that 's just another phrasing of being unfair . i decided to think for myself logically why being unfair is wrong and so far i 've only managed to come up with a few flawed answers . firstly , being unfair is self-evidently wrong . now this works out , until you realise that different people in different environments would find different values to be self-evident . \textcolor{lightblue}{\em for example , if a human were to grow up alone without being in contact with another human since he was first conscious , then what he would n't find fairness to be self-evident} . instead , what he wo n't hesitate to do is kill others for whatever reasons he sees fit . he would see what he does as being acceptable , but we would n't . however , it would be impossible to convince him that others have a right to life because he grew not knowing empathy . if we apply this to the current context , then people in the west find different morals to be self-evident than people in asia or the middle east . yet everyone claims the other is inhumane , with no explanation how it is inhumane , or what is inhumane . another answer why being unfair is wrong is that it without fairness , society would n't function optimally . however if i purge the retired elderly or the ill who needlessly consume resources , then it would boost the cogs of society , wouldnt it ? its still considered wrong . therefore this answer is invalid . anyone have answers for the question " why is being unfair wrong ? "
		 \\
		\midrule
		{\bf Our Counter:} how is being unfair any better than being wrong? fairness is subjective. in any society, fairness is subjective. if a person has a problem, does that mean their position is fair? the way we live the consequences of their decision means we can't change them. but why is that wrong?  \\
		\midrule
		{\bf Hua and Wang's Counter:} i think it 's important to distinguish between the two scenarios , and i think that it 's more important to understand what you mean by " different " . i think you 're correct , but i think it \\
		\bottomrule
	\end{tabular} 
	\caption{A list of examples of counter-arguments generated by our approach and by the approach of Hua and Wang (2019). The italicized premise segment was identified as the weak premise by our approach.}
	\label{table-generation-examples-full}
\end{figure*}
	
\end{document}